\documentclass[twoside]{article}
\usepackage{amsmath,amsthm,verbatim,amssymb,amsfonts,amscd, graphicx}
\usepackage{graphics}

\newcommand{\bv}{{\boldsymbol b}}
\newcommand{\cv}{{\boldsymbol c}}
\newcommand{\dv}{{\boldsymbol d}}

\newcommand{\fv}{{\boldsymbol f}}

\newcommand{\hv}{{\boldsymbol h}}
\newcommand{\iv}{{\boldsymbol i}}

\newcommand{\ov}{{\boldsymbol o}}

\newcommand{\tv}{{\boldsymbol t}}

\newcommand{\Wv}{{\bf W}}
\newcommand{\Vv}{{\bf V}}

\newcommand{\Uv}{{\bf U}}
\newcommand{\xv}{{\boldsymbol x}}

\newcommand{\betav}{{\boldsymbol \beta}}
\newcommand{\thetav}{{\boldsymbol \theta}}

\newcommand{\R}{\mathbb{R}}
\newcommand{\Wcal}{\mathcal{W}}
\newcommand{\Ucal}{\mathcal{U}}

\usepackage[accepted]{aistats2018}
\usepackage{multirow}
\usepackage{tabularx}
\usepackage{ctable}
\usepackage{natbib}
\usepackage{arydshln}
\bibpunct{(}{)}{;}{a}{,}{,}
\usepackage{indentfirst}

\begin{document}
	
	%
	\runningtitle{Topic Compositional Neural Language Model}
	
	%
	\runningauthor{W. Wang, Z. Gan, W. Wang, D. Shen, J. Huang, W. Ping, S. Satheesh, L. Carin}
	
	\twocolumn[

	\aistatstitle{Topic Compositional Neural Language Model}

	\aistatsauthor{Wenlin Wang$^{1}$ \And Zhe Gan$^{1}$ \And  Wenqi Wang$^{3}$ \And Dinghan Shen$^{1}$ \And Jiaji Huang$^{2}$}
	\aistatsauthor{Wei Ping$^{2}$ \And  Sanjeev Satheesh$^{2}$ \And Lawrence Carin$^{1}$}

	\aistatsaddress{$^{1}$Duke University \And  $^{2}$Baidu Silicon Valley AI Lab \And $^{3}$Purdue University}]
	
	\begin{abstract}
		We propose a Topic Compositional Neural Language Model (TCNLM), a novel method designed to simultaneously capture both the \emph{global} semantic meaning and the \emph{local} word-ordering structure in a document. The TCNLM learns the global semantic coherence of a document via a neural topic model, and the probability of each learned latent topic is further used to build a Mixture-of-Experts (MoE) language model, where each expert (corresponding to one topic) is a recurrent neural network (RNN) that accounts for learning the local structure of a word sequence. In order to train the MoE model efficiently, a matrix factorization method is applied, by extending each weight matrix of the RNN to be an ensemble of topic-dependent weight matrices. The degree to which each member of the ensemble is used is tied to the document-dependent probability of the corresponding topics. Experimental results on several corpora show that the proposed approach outperforms both a pure RNN-based model and other topic-guided language models. Further, our model yields sensible topics, and also has the capacity to generate meaningful sentences conditioned on given topics. 
	\end{abstract}
	
	\section{Introduction}
	
	\begin{figure*}
		\centering
		\includegraphics[scale=0.35]{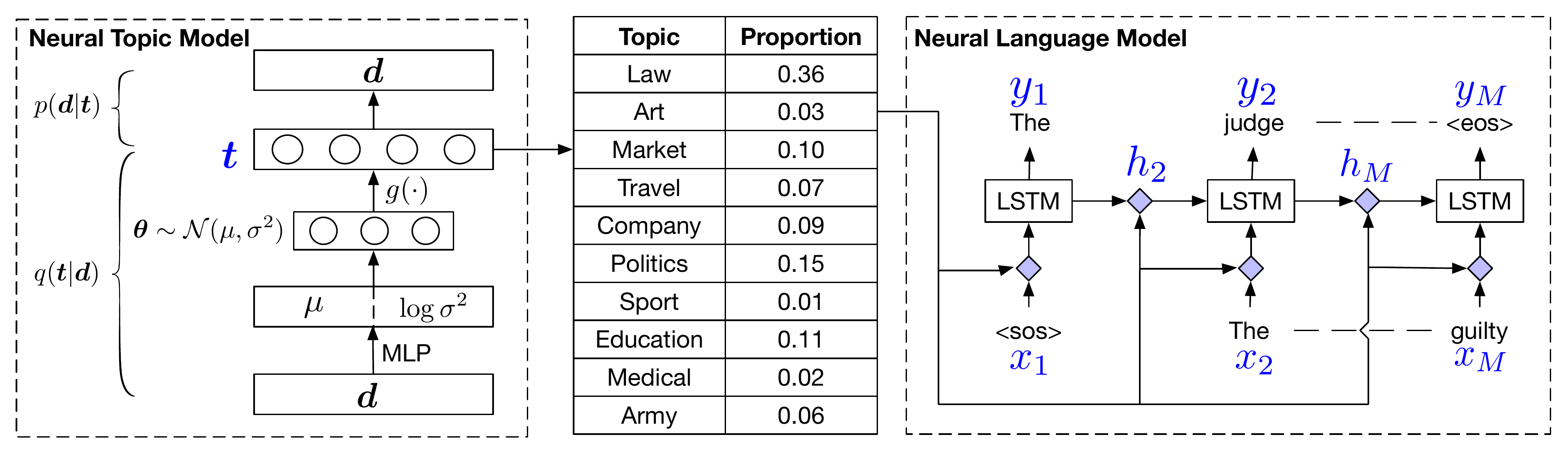}
		\label{fig:architecture}
		\caption{The overall architecture of the proposed model.}
	\end{figure*}
	
	A language model is a fundamental component to natural language processing (NLP). It plays a key role in many traditional NLP tasks, ranging from speech recognition~\citep{mikolov2010recurrent,arisoy2012deep,sriram2017cold}, machine translation~\citep{schwenk2012large,vaswani2013decoding} to image captioning~\citep{mao2014deep,devlin2015language}. Training a good language model often improves the underlying metrics of these applications, \textit{e.g.,} word error rates for speech recognition and BLEU scores~\citep{papineni2002bleu} for machine translation. Hence, learning a powerful language model has become a central task in NLP. Typically, the primary goal of a language model is to predict distributions over words, which has to encode both the semantic knowledge and grammatical structure in the documents. RNN-based neural language models have yielded state-of-the-art performance~\citep{jozefowicz2016exploring,shazeer2017outrageously}. However, they are typically applied only at the sentence level, without access to the broad document context. Such models may consequently fail to capture long-term dependencies of a document~\citep{dieng2016topicrnn}.
	
	Fortunately, such broader context information is of a semantic nature, and can be captured by a topic model. Topic models have been studied for decades and have become a powerful tool for extracting high-level semantic structure of document collections, by inferring latent topics. The classical Latent Dirichlet Allocation (LDA) method~\citep{blei2003latent} and its variants, including recent work on neural topic models~\citep{wan2012hybrid,cao2015novel,miao2017discovering}, have been useful for a plethora of applications in NLP. 
	
	Although language models that leverage topics have shown promise, they also have several limitations. For example, some of the existing methods use only pre-trained topic models~\citep{mikolov2012context}, without considering the word-sequence prediction task of interest. Another key limitation of the existing methods lies in the integration of the learned topics into the language model; \emph{e.g.}, either through concatenating the topic vector as an additional feature of RNNs~\citep{mikolov2012context,lau2017topically}, or re-scoring the predicted distribution over words using the topic vector~\citep{dieng2016topicrnn}. The former requires a balance between the number of RNN hidden units and the number of topics, while the latter has to carefully design the vocabulary of the topic model. 
	
	Motivated by the aforementioned goals and limitations of existing approaches, we propose the Topic Compositional Neural Language Model (TCNLM), a new approach to simultaneously learn a neural topic model and a neural language model. As depicted in Figure~\ref{fig:architecture}, TCNLM learns the latent topics within a variational autoencoder~\citep{kingma2013auto} framework, and the designed latent code $\tv$ quantifies the probability of topic usage within a document. Latent code $\tv$ is further used in a Mixture-of-Experts model~\citep{hu1997patient}, where each latent topic has a corresponding language model (expert). A combination of these ``experts,'' weighted by the topic-usage probabilities, results in our prediction for the sentences. A matrix factorization approach is further utilized to reduce computational cost as well as prevent overfitting. The entire model is trained end-to-end by maximizing the variational lower bound.
	Through a comprehensive set of experiments, we demonstrate that the proposed model is able to significantly reduce the perplexity of a language model and effectively assemble the meaning of topics to generate meaningful sentences. Both quantitative and qualitative comparisons are provided to verify the superiority of our model. 
	
	\section{Preliminaries}
	We briefly review RNN-based language models and traditional probabilistic topic models.
	
	\paragraph{Language Model}
	A language model aims to learn a probability distribution over a sequence of words in a pre-defined vocabulary. We denote $\mathcal{V}$ as the vocabulary set and $\{y_1, ..., y_M\}$ to be a sequence of words, with each $y_m\in \mathcal{V}$. A language model defines the likelihood of the sequence through a joint probability distribution
	\begin{align}
	p(y_1, ..., y_M) = p(y_1)\prod_{m=2}^{M}p(y_m | y_{1:m-1})\,.
	\end{align}
	
	RNN-based language models define the conditional probabiltiy of each word $y_m$ given all the previous words $y_{1:m-1}$ through the hidden state $\hv_m$: 
	\begin{align}
	p(y_m | y_{1:m-1}) &= p(y_m | \hv_m)  \label{lm}\\
	\hv_m &= f(\hv_{m-1}, x_m) \,. 
	\end{align}
	The function $f(\cdot)$ is typically implemented as a basic RNN cell, a Long Short-Term Memory (LSTM) cell~\citep{hochreiter1997long}, or a Gated Recurrent Unit (GRU) cell~\citep{cho2014learning}. The input and output words are related via the relation $x_m = y_{m-1}$. 
	
	\paragraph{Topic Model}
	A topic model is a probabilistic graphical representation for uncovering the underlying semantic structure of a document collection. Latent Dirichlet Allocation (LDA)~\citep{blei2003latent}, for example, provides a robust and scalable approach for document modeling, by introducing latent variables for each token, indicating its topic assignment. Specifically, let $\tv$ denote the topic proportion for document $d$, and $z_n$ represent the topic assignment for word $w_n$. The Dirichlet distribution is employed as the prior of $\tv$. The generative process of LDA may be summarized as:
	\begin{align*}
	\tv  \sim \text{Dir}(\alpha_0), z_n \sim \text{Discrete}  (\tv)\,,
	w_n \sim \text{Discrete} (\betav_{z_n}) \,,
	\end{align*}
	where $\betav_{z_n}$ represents the distribution over words for topic $z_n$, $\alpha_0$ is the hyper-parameter of the Dirichlet prior, $n\in [1, N_d ]$, and $N_d$ is the number of words in document $d$. The marginal likelihood for document $d$ can be expressed as 
	\begin{align*}
	p(d | \alpha_0, \betav) = \int_{\tv}  p(\tv | \alpha_0) \prod_n \sum_{z_n} p(w_n | \betav_{z_n})p(z_n | \tv) d\tv \,.
	\end{align*}
	
	\section{Topic Compositional Neural Language Model}
	
	We describe the proposed TCNLM, as illustrated in Figure~\ref{fig:architecture}. Our model consists of two key components: (\emph{i}) a neural topic model (NTM), and (\emph{ii}) a neural language model (NLM). The NTM aims to capture the long-range semantic meanings across the document, while the NLM is designed to learn the local semantic and syntactic relationships between words. 
	\subsection{Neural Topic Model}
	Let $\dv\in \mathbb{Z}_+^D$ denote the bag-of-words representation of a document, with $\mathbb{Z}_+$ denoting nonnegative integers. $D$ is the vocabulary size, and each element of $\dv$ reflects a count of the number of times the corresponding word occurs in the document. Distinct from LDA~\citep{blei2003latent}, we pass a \emph{Gaussian} random vector through a softmax function to parameterize the multinomial document topic distributions~\citep{miao2017discovering}. Specifically, the generative process of the NTM is 
	\begin{align}
	\thetav &\sim \mathcal{N}(\mu_0, \sigma_0^2) & \tv &= g(\thetav) \nonumber\\
	z_n &\sim \text{Discrete}(\tv) & w_n &\sim \text{Discrete}(\betav_{z_n})\,,
	\end{align}
	where $\mathcal{N}(\mu_0, \sigma_0^2)$ is an isotropic Gaussian distribution, with mean $\mu_0$ and variance $\sigma_0^2$ in each dimension; $g(\cdot)$ is a transformation function that maps sample $\thetav$ to the topic embedding $\tv$, defined here as $g(\thetav) = \mbox{softmax}(\hat{\Wv}\thetav + \hat{\bv})$, where $\hat{\Wv}$ and $\hat{\bv}$ are trainable parameters. 
	
	The marginal likelihood for document $\dv$ is:
	{\small
		\begin{align}
		p(\dv | \mu_0, \sigma_0, \betav) &= \int_{\tv}p(\tv | \mu_0, \sigma_0^2)  \prod_n \sum_{z_n}p(w_n | \betav_{z_n}) p(z_n | \tv) d\tv \nonumber \\
		&= \int_{\tv}p(\tv | \mu_0, \sigma_0^2) \prod_n p(w_n | \betav, \tv) d\tv \nonumber \\
		& =  \int_{\tv}p(\tv | \mu_0, \sigma_0^2)  p(\dv | \betav, \tv) d\tv \,.
		\label{eq1}
		\end{align}
	}
	\noindent The second equation in~(\ref{eq1}) holds because we can readily marginalized out the sampled topic words $z_n$ by 
	\begin{align}
	p(w_n | \betav, \tv ) =  \sum_{z_n}p(w_n | \betav_{z_n}) p(z_n | \tv)  = \betav\tv \,.
	\end{align}
	$\betav = \{\betav_1, \betav_2, ..., \betav_T\}$ is the transition matrix from the topic distribution to the word distribution, which are trainable parameters of the decoder; $T$ is the number of topics and $\betav_i\in \R^D$ is the topic distribution over words (all elements of $\betav_i$ are nonnegative, and they sum to one). 
	
	The re-parameterization trick~\citep{kingma2013auto} can be applied to build an unbiased and low-variance gradient estimator for the variational distribution. The parameter updates can still be derived directly from the variational lower bound, as discussed in Section~\ref{sec:model_inference}.
	
	\paragraph{Diversity Regularizer} Redundance in inferred topics is a common issue exisiting in general topic models. In order to address this issue, it is straightforward to regularize the row-wise distance between each paired topics to diversify the topics. Following~\citet{xie2015diversifying,miao2017discovering}, we apply a topic diversity regularization while carrying out the inference. 
	
	Specifically, the distance between a pair of topics are measured by their cosine distance $a(\betav_i, \betav_j) = \arccos \left( \frac{|\betav_i\cdot \betav_j|}{||\betav_i||_2 ||\betav_j||_2}\right)$. The mean angle of all pairs of $T$ topics is $\phi = \frac{1}{T^2}\sum_i\sum_j a(\betav_i, \betav_j)$, and the variance is $\nu = \frac{1}{T^2}\sum_i\sum_j(a(\betav_i, \betav_j) - \phi)^2$. Finally, the topic diversity regularization is defined as $R=\phi - \nu$. 
	
	\subsection{Neural Language Model}
	
	We propose a Mixture-of-Experts (MoE) language model, which consists a set of ``expert networks'', \emph{i.e.}, $E_1, E_2, ..., E_T$. Each expert is itself an RNN with its own parameters corresponding to a latent topic. 
	
	Without loss of generality, we begin by discussing an RNN with a simple transition function, which is then generalized to the LSTM. Specifically, we define two weight tensors $\Wcal \in \R^{n_h\times n_x \times T}$ and $\Ucal \in \R^{n_h\times n_h \times T}$, where $n_h$ is the number of hidden units and $n_x$ is the dimension of word embedding. Each expert $E_k$ corresponds to a set of parameters $\Wcal[k]$ and $\Ucal[k]$, which denotes the $k$-th 2D ``slice'' of $\Wcal$ and $\Ucal$, respectively. All $T$ experts work cooperatively to generate an output $y_m$. Sepcifically,  
	\begin{align}
	p(y_m) &= \sum_{k=1}^T \tv_k \cdot \mbox{softmax}(\Vv \hv_m^{(k)}) \label{eq:MoE} \\
	\hv_m^{(k)} &= \sigma(\Wcal[k]\xv_m + \Ucal[k]\hv_{m-1}) \,,
	\end{align}
	where  $\tv_k$ is the usage of topic $k$ (component $k$ of $\tv$), and $\sigma(\cdot)$ is a sigmoid function; $\Vv$ is the weight matrix connecting the RNN's hidden state, used for computing a
	distribution over words. Bias terms are omitted for simplicity.
	
	However, such an MoE module is computationally prohibitive and storage excessive. The training process is inefficient and even infeasible in practice. To remedy this, instead of ensembling the output of the $T$ experts as in (\ref{eq:MoE}), we extend the weight matrix of the RNN to be an ensemble of topic-dependent weight matrices. Specifically, the $T$ experts work together as follows:
	\begin{align}
	p(y_m) &=\mbox{softmax}(\Vv \hv_m)  \\
	\hv_m &= \sigma(\Wv(\tv) \xv_m + \Uv(\tv) \hv_{m-1}) \,,
	\end{align}
	and 
	\begin{align}
	\Wv(\tv) =  \sum_{k=1}^T \tv_k\cdot\Wcal[k], \,\, \Uv(\tv) = \sum_{k=1}^T \tv_k \cdot\Ucal[k] \label{eq:tensor}\,.
	\end{align}
	
	In order to reduce the number of model parameters, motivated by~\citet{gan2016semantic,song2016factored}, instead of implementing a tensor as in (\ref{eq:tensor}), we decompose $\Wv(\tv)$ into a multiplication of three terms $\Wv_a\in R^{n_h\times n_f}$, $\Wv_b\in R^{n_f\times T}$ and $\Wv_c\in R^{n_f\times n_x}$, where $n_f$ is the number of factors. Specifically, 
	\begin{align}
	\Wv(\tv) &= \Wv_a \cdot \text{diag}(\Wv_{b}\tv) \cdot \Wv_{c} \nonumber \\
	&= \Wv_a \cdot (\Wv_b\tv\odot \Wv_c) \,,
	\end{align}
	where $\odot$ represents the Hadamard operator. $\Wv_a$ and $\Wv_c$ are shared parameters across all topics, to capture the common linguistic patterns. $\Wv_b$ are the factors which are weighted by the learned topic embedding $\tv$. The same factorization is also applied for $\Uv(\tv)$. The topic distribution $\tv$ affects RNN parameters associated with the document when predicting the succeeding words, which implicitly defines an ensemble of $T$ language models.
	In this factorized model, the RNN weight matrices that correspond to each topic
	share ``structure''. 
	
	Now we generalize the above analysis by using LSTM units. Specifically, we summarize the new topic compositional LSTM cell as:
	\begin{align}
	\iv_m   &= \sigma(\Wv_{ia}\tilde{\xv}_{i, m-1} + \Uv_{ia}\tilde{\hv}_{i, m-1}) \nonumber\\
	\fv_m  &= \sigma(\Wv_{fa}\tilde{\xv}_{f, m-1} + \Uv_{fa}\tilde{\hv}_{f, m-1}) \nonumber\\
	\ov_m &= \sigma(\Wv_{oa}\tilde{\xv}_{o, m-1} +\Uv_{oa}\tilde{\hv}_{o, m-1}) \nonumber\\
	\tilde{\cv}_m &= \sigma(\Wv_{ca}\tilde{\xv}_{c, m-1} +\Uv_{ca}\tilde{\hv}_{c, m-1}) \nonumber\\
	\cv_m &= \iv_m\odot \tilde{\cv}_m  + \fv_m \cdot \cv_{m-1} \nonumber\\
	\hv_m &= \ov_m \odot \text{tanh} (\cv_m) \,.
	\end{align}
	For $*= i, f, o, c$, we define 
	\begin{align}
	\tilde{\xv}_{\ast, m-1} &= \Wv_{\ast b}\tv\odot\Wv_{\ast c}\xv_{m-1} \label{eq:compute_x}\\
	\tilde{\hv}_{\ast, m-1} &= \Uv_{\ast b}\tv\odot\Uv_{\ast c}\hv_{m-1} \label{eq:compute_h} \,.
	\end{align}
	Compared with a standard LSTM cell, our LSTM unit has a total number of parameters in size of $4n_f\cdot (n_x + 2T + 3n_h)$ and the additional computational cost comes from~(\ref{eq:compute_x}) and~(\ref{eq:compute_h}). Further, empirical comparison has been conducted in Section 5.6 to verify that our proposed model is superior than using the naive MoE implementation as in~(\ref{eq:MoE}). 
	
	\subsection{Model Inference} \label{sec:model_inference}
	
	The proposed model (see Figure~\ref{fig:architecture}) follows the variational autoencoder~\citep{kingma2013auto} framework, which takes the bag-of-words as input and embeds a document into the topic vector. This vector is then used to reconstruct the bag-of-words input, and also to learn an ensemble of RNNs for predicting a sequence of words in the document. 
	
	The joint marginal likelihood can be written as:
	\begin{align}
	p(y_{1:M}, \dv |\mu_0, \sigma_0^2, \betav) &= \int_\tv p(\tv | \mu_0, \sigma^2_0) p(\dv | \betav, \tv) \nonumber \\
	&\prod_{m=1}^{M}   p(y_m | y_{1:m-1}, \tv)  d\tv \,.   \label{eq:joint_likelihood}
	\end{align}
	Since the direct optimization of~(\ref{eq:joint_likelihood}) is intractable, we employ variational inference~\citep{jordan1999introduction}. We denote $q(\tv | \dv)$ to be the variational distribution for $\tv$. Hence, we construct the variational objective function, also called the evidence lower bound (ELBO), as 
	\begin{align}
	\mathcal{L}&=  \underbrace{ \mathbb{E}_{q(\tv | \dv)} \left( \log p( \dv | \tv) \right)- \text{KL}\left( q(\tv | \dv) || p(\tv | \mu_0, \sigma_0^2) \right) }_{\text{neural topic model}} \nonumber\\
	&+  \underbrace{\mathbb{E}_{q(\tv | \dv)} \left( \sum_{m=1}^{M}\log p(y_m | y_{1:m-1}, \tv) \right)}_{\text{neural language model}} \\
	& \leq \log p(y_{1:M}, \dv | \mu_0, \sigma_0^2, \betav) \nonumber \,.
	\end{align}
	More details can be found in the Supplementary Material.
	In experiments, we optimize the ELBO together with the diversity regularisation:
	\begin{align}
	\mathcal{J} = \mathcal{L} + \lambda \cdot R \,.
	\end{align}
	
	\section{Related Work}\label{sec:related}
	
		\begin{table*}[!htbp]
		\begin{center}
			\scalebox{0.83}{
				\begin{tabular}{l|c c|c  c c| c  c c| c c c}
					\specialrule{.1em}{.05em}{.05em} 
					\multirow{2}{*}{\textbf{Dataset}} &
					\multicolumn{2}{c|}{\textbf{Vocabulary}} & \multicolumn{3}{c|}{\textbf{Training}} & \multicolumn{3}{c|}{\textbf{Development}} & \multicolumn{3}{c}{\textbf{Testing}} \\
					& LM & TM & \# Docs & \# Sents & \# Tokens & \# Docs & \# Sents & \# Tokens & \# Docs & \# Sents & \# Tokens \\
					\hline
					\textsc{APNEWS} & $32,400$ & $7,790$& $50K$ & $0.7M$ & $15M$ & $2K$ & $27.4K$ & $0.6M$ & $2K$ & $26.3K$ & $0.6M$ \\ \hline
					\textsc{IMDB} & $34,256$ & $8,713$& $75K$ & $0.9M$ & $20M$ & $12.5K$ & $0.2M$ & $0.3M$ & $12.5K$ & $0.2M$ & $0.3M$ \\ \hline
					\textsc{BNC} & $41,370$ & $9,741$& $15K$ & $0.8M$ & $18M$ & $1K$ & $44K$ & $1M$ & $1K$ & $52K$ & $1M$ \\ 
					\specialrule{.1em}{.05em}{.05em} 
			\end{tabular}}
			\caption{Summary statistics for the datasets used in the experiments.}
			\label{Table:datasets}
		\end{center}
		\vspace{-4pt}
	\end{table*} 
	
	\textbf{Topic Model } Topic models have been studied for a variety of applications in document modeling. Beyond LDA~\citep{blei2003latent}, significant extensions have been proposed, including capturing topic correlations~\citep{blei2007correlated}, modeling temporal dependencies~\citep{blei2006dynamic}, discovering an unbounded number of topics~\citep{teh2005sharing}, learning deep architectures~\citep{henao2015deep,zhou2015poisson}, among many others. Recently, neural topic models have attracted much attention, building upon the successful usage of restricted Boltzmann machines~\citep{hinton2009replicated}, auto-regressive models~\citep{larochelle2012neural}, sigmoid belief networks~\citep{gan2015scalable}, and variational autoencoders~\citep{miao2016neural}.
	
	Variational inference has been successfully applied in a variety of applications~\citep{pu2016variational,wang2017zero, chen2017continuous}. The recent work of~\citet{miao2017discovering} employs variational inference to train topic models, and is closely related to our work. Their model follows the original LDA formulation and extends it by parameterizing the multinomial distribution with neural networks. In contrast, our model enforces the neural network not only modeling documents as bag-of-words, but also transfering the inferred topic knowledge to a language model for word-sequence generation. 
	
	\paragraph{Language Model} Neural language models have recently achieved remarkable advances~\citep{mikolov2010recurrent}. The RNN-based language model (RNNLM) is superior for its ability to model longer-term temporal dependencies without imposing a strong conditional independence assumption; it has recently been shown to outperform carefully-tuned traditional n-gram-based language models~\citep{jozefowicz2016exploring}. 
	
An	RNNLM can be further improved by utilizing the broad document context~\citep{mikolov2012context}. Such models typically extract latent topics via a topic model, and then send the topic vector to a language model for sentence generation. Important work in this direction include~\citet{mikolov2012context, dieng2016topicrnn, lau2017topically, ahn2016neural}. The key differences of these methods is in either the topic model itself or the method of integrating the topic vector into the language model. In terms of the topic model,~\citet{mikolov2012context} uses a pre-trained LDA model;~\citet{dieng2016topicrnn} uses a variational autoencoder;~\citet{lau2017topically} introduces an attention-based convolutional neural network to extract semantic topics; and~\citet{ahn2016neural} utilizes the topic associated to the fact pairs derived from a knowledge graph~\citep{vinyals2015neural}. 
	
Concerning the method of incorporating the topic vector into the language model,~\citet{mikolov2012context} and~\citet{lau2017topically} extend the RNN cell with additional topic features.~\citet{dieng2016topicrnn} and~\citet{ahn2016neural} use a hybrid model combining the predicted word distribution given by both a topic model and a standard RNNLM. Distinct from these approaches, our model learns the topic model and the language model jointly under the VAE framework, allowing an efficient end-to-end training process. Further, the topic information is used as guidance for a Mixture-of-Experts (MoE) model design. Under our factorization method, the model can yield boosted performance efficiently (as corroborated in the experiments).  
	
	Recently,~\citet{shazeer2017outrageously} proposes a MoE model for large-scale language modeling. Different from ours, they introduce a MoE layer, in which each expert stands for a small feed-forward neural network on the previous output of the LSTM layer. Therefore, it yields a significant quantity of additional parameters and computational cost, which is infeasible to train on a single GPU machine. Moreover, they provide no semantic meanings for each expert, and all experts are treated equally; the proposed model can generate meaningful sentences conditioned on given topics. 
	
	Our TCNLM is similar to~\citet{gan2016semantic}. However, \citet{gan2016semantic} uses a two-step pipline, first learning a multi-label classifier on a group of pre-defined image tags, and then generating image captions conditioned on them. In comparison, our model jointly learns a topic model and a language model, and focuses on the language modeling task.
	
	\section{Experiments}\label{sec:exp}
	
	\begin{table*}[!htbp]
		\begin{center}
			\scalebox{0.7}{
				\begin{tabular}{c |c| c | c c c | c | c c c | c c c | c c c}
					\specialrule{.1em}{.05em}{.05em} 
					\multirow{2}{*}{\textbf{Dataset}} & \textbf{LSTM} & \multirow{2}{*}{\textbf{basic-LSTM$^*$}} &\multicolumn{3}{c|}{\textbf{LDA+LSTM}$^*$} &\multirow{2}{*}{\textbf{LCLM}$^*$} &\multicolumn{3}{c|}{\textbf{Topic-RNN}} &\multicolumn{3}{c|}{\textbf{TDLM}$^*$} 
					&\multicolumn{3}{c}{\textbf{TCNLM}} \\ 
					& \textbf{type}& & 50 & 100 & 150 & & 50 & 100 & 150 & 50 & 100 & 150 & 50 & 100 & 150\\ \hline
					\multirow{2}{*}{\textsc{APNEWS}} & small & $64.13$& $57.05$ & $55.52$ & $54.83$ & $54.18$ &$56.77$ &$54.54$ &$54.12$ &$53.00$ &$52.75$ &$52.65$ & $52.75$  &$52.63$ & $\textbf{52.59}$ \\
					& large & $58.89$ & $52.72$ & $50.75$ & $50.17$ & $50.63$ &$53.19$ &$50.24$ &$50.01$ & $48.96$ &$48.97$ & $48.21$ & $48.07$  &$47.81$ &$\textbf{47.74}$ \\ \hline
					\multirow{2}{*}{\textsc{IMDB}} & small & $72.14$& $69.58$ & $69.64$ & $69.62$ & $67.78$ &$68.74$ &$67.83$ &$66.45$ &$63.67$ &$63.45$ &$63.82$ & $63.98$ &$62.64$ & $\textbf{62.59}$ \\
					& large & $66.47$ & $63.48$ & $63.04$ & $62.78$ & $67.86$ &$63.02$ &$61.59$ &$60.14$ &$58.99$ & $59.04$ & $58.59$ & $57.06$  &$56.38$ &$\textbf{56.12}$  \\ \hline
					\multirow{2}{*}{\textsc{BNC}} & small & $102.89$& $96.42$ & $96.50$ & $96.38$ & $87.47$ &$94.66$ &$93.57$ &$93.55$ & $87.42$ & $85.99$ & $86.43$ & $87.98$ & $86.44$ & $\textbf{86.21}$ \\
					& large & $94.23$ & $88.42$ & $87.77$ & $87.28$ & $80.68$&$85.90$ &$84.62$ &$84.12$ & $82.62$ & $81.83$ & $80.58$ & $80.29$ &$80.14$ & $\textbf{80.12}$ \\
					\specialrule{.1em}{.05em}{.05em} 
			\end{tabular}}
			\caption{Test perplexities of different models on \textsc{APNEWS}, \textsc{IMDB} and \textsc{BNC}. ($*$) taken from~\citet{lau2017topically}.}
			\label{Table:resLM}
			\vspace{-10pt}
		\end{center}
	\end{table*} 
	\paragraph{Datasets} We present experimental results on three publicly available corpora: \textsc{APNEWS}, \textsc{IMDB} and \textsc{BNC}. 
	\textsc{APNEWS}\footnote{https://www.ap.org/en-gb/} is a collection of Associated Press news articles from 2009 to 2016. \textsc{IMDB} is a set of movie reviews collected by~\citet{maas2011learning}, and \textsc{BNC}~\citep{BNCConsortium2007} is the written portion of the British National Corpus, which contains excerpts from journals, books, letters, essays, memoranda, news and other types of text. These three datasets can be downloaded from GitHub\footnote{https://github.com/jhlau/topically-driven-language-model}. 
	
	We follow the preprocessing steps in~\citet{lau2017topically}. Specifically, words and sentences are tokenized using Stanford CoreNLP~\citep{manning2014stanford}. We lowercase all word tokens, and filter out word tokens that occur less than 10 times. For topic modeling, we additionally remove stopwords\footnote{We use the following stopwords list: https://github.\\com/mimno/Mallet/blob/master/stoplists/en.txt} in the documents and exclude the top $0.1\%$ most frequent words and also words that appear in less than 100 documents. All these datasets are divided into training, development and testing sets. A summary statistic of these datasets is provided in Table~\ref{Table:datasets}.
	
	\paragraph{Setup} For the NTM part, we consider a 2-layer feed-forward neural network to model $q(\tv | \dv)$, with $256$ hidden units in each layer; ReLU~\citep{nair2010rectified} is used as the activation function. The hyper-parameter $\lambda$ for the diversity regularizer is fixed to 0.1 across all the experiments. All the sentences in a paragraph, excluding the one being predicted, are used to obtain the bag-of-words document representation $\dv$. The maximum number of words in a paragraph is set to $300$. 
	
	In terms of the NLM part, we consider 2 settings: (\emph{i}) a small 1-layer LSTM model with $600$ hidden units, and (\emph{ii}) a large 2-layer LSTM model with $900$ hidden units in each layer. The sequence length is fixed to 30.  In order to alleviate overfitting, dropout with a rate of $0.4$ is used in each LSTM layer. In addition, adaptive softmax~\citep{grave2016efficient} is used to speed up the training process.
	
	During training, the NTM and NLM parameters are jointly learned using Adam~\citep{kingma2014adam}. All the hyper-parameters are tuned based on the performance on the development set. We empirically find that the optimal settings are fairly robust across the 3 datasets. All the experiments were conducted using Tensorflow and trained on NVIDIA GTX TITAN X with 3072 cores and 12GB global memory.
	
	\subsection{Language Model Evaluation}
	
	Perplexity is used as the metric to evaluate the performance of the language model. In order to demonstrate the advantage of the proposed model, we compare TCNLM with the following baselines:
	\begin{itemize}
		\item \textbf{basic-LSTM}: A baseline LSTM-based language model, using the same architecture and hyper-parameters as TCNLM wherever applicable.
		\item \textbf{LDA+LSTM}: A topic-enrolled LSTM-based language model. We first pretrain an LDA model~\citep{blei2003latent} to learn 50/100/150 topics for \textsc{APNEWS}, \textsc{IMDB} and \textsc{BNC}. Given a document, the LDA topic distribution is incorporated by concatenating with the output of the hidden states to predict the next word. 
		\item \textbf{LCLM}~\citep{Wang2016lLarger}: A context-based language model, which incorporates context information from preceding sentences. The preceding sentences are treated as bag-of-words, and an attention mechanism is used when predicting the next word. 
		\item \textbf{TDLM}~\citep{lau2017topically}: A convolutional topic model enrolled languge model. Its topic knowledge is utilized by concatenating to a dense layer of a recurrent language model. 
		\item \textbf{Topic-RNN}~\citep{dieng2016topicrnn}: A joint learning framework that learns a topic model and a language model simutaneously. The topic information is incorporated through a linear transformation to rescore the prediction of the next word. 
	\end{itemize}
	
	Topic-RNN (\citealp{dieng2016topicrnn}) is implemented by ourselves and other comparisons are copied from \citep{lau2017topically}. Results are presented in Table~\ref{Table:resLM}. We highlight some observations.
	(\textit{i}) All the topic-enrolled methods outperform the basic-LSTM model, indicating the effectiveness of incorporating global semantic topic information. (\textit{ii}) Our TCNLM performs the best across all datasets, and the trend keeps improving with the increase of topic numbers. (\textit{iii}) The improved performance of TCNLM over LCLM implies that encoding the document context into meaningful topics provides a better way to improve the language model compared with using the extra context words directly. (\textit{iv}) The margin between LDA+LSTM/Topic-RNN and our TCNLM indicates that our model supplies a more efficient way to utilize the topic information through the joint variational learning framework to implicitly train an ensemble model.
	
	\begin{table*}[!htbp]
		\begin{center}
			\scalebox{0.65}{
				\begin{tabular}{l | c c c c c c c c c c }
					\specialrule{.1em}{.05em}{.05em} 
					\textbf{Dataset} &  \textbf{army} & \textbf{animal} & \textbf{medical} & \textbf{market} & \textbf{lottory} &\textbf{terrorism} &  \textbf{law} & \textbf{art} &\textbf{transportation} & \textbf{education} \\ \hline
					\multirow{5}{*}{\textsc{APNEWS}} & afghanistan&animals &patients &zacks &casino &syria &lawsuit &album & airlines &students\\
					&veterans &dogs &drug &cents &mega &iran &damages &music & fraud &math \\ 
					&soldiers &zoo &fda &earnings &lottery &militants  &plaintiffs &film &scheme &schools \\
					&brigade &bear &disease &keywords &gambling &al-qaida  &filed &songs & conspiracy & education\\
					&infantry &wildlife &virus &share &jackpot &korea  &suit &comedy &flights &teachers \\ \hline
					\multirow{6}{*}{\textsc{IMDB}} & \textbf{horror}&\textbf{action} &\textbf{family} &\textbf{children} &\textbf{war} &\textbf{detective} &\textbf{sci-fi} &\textbf{negative} &\textbf{ethic} &\textbf{epsiode} \\ \cline{2-11}
					& zombie& martial& rampling& kids& war& eyre& alien& awful& gay& season\\ 
					& slasher& kung& relationship& snoopy& che& rochester& godzilla& unfunny& school& episodes\\
					& massacre& li& binoche& santa& documentary& book& tarzan& sex&girls & series \\
					& chainsaw& chan& marie& cartoon& muslims& austen& planet& poor& women&  columbo\\
					& gore& fu& mother& parents& jews& holmes& aliens& worst& sex& batman\\ \hline
					\multirow{6}{*}{\textsc{BNC}} & \textbf{environment} & \textbf{education} & \textbf{politics} & \textbf{business} & \textbf{facilities} & \textbf{sports} & \textbf{art} & \textbf{award} & \textbf{expression} & \textbf{crime} \\ \cline{2-11}
					&pollution &courses &elections &corp & bedrooms& goal &album &john &eye &police\\
					&emissions &training &economic &turnover &hotel &score &band &award &looked &murder \\
					&nuclear &students &minister &unix &garden &cup &guitar &research &hair &killed \\
					&waste &medau &political &net &situated &ball &music &darlington &lips &jury \\
					&environmental &education &democratic &profits &rooms &season &film &speaker &stared &trail \\ \hline
			\end{tabular}}
			\label{Table:topicwords}
			\vspace{-.5em}
			\caption{10 topics learned from our TCNLM on \textsc{APNEWS}, \textsc{IMDB} and \textsc{BNC}.}
			\vspace{-2em}
		\end{center}
	\end{table*} 
	
	\begin{table}[!htbp]
		\begin{center}
			\scalebox{0.66}{
				\begin{tabular}{c  c c c c  }
					\specialrule{.1em}{.05em}{.05em} 
					\multirow{2}{*}{\textbf{\# Topic}} & \multirow{2}{*}{\textbf{Model}} &  \multicolumn{3}{c}{\textbf{Coherence}} \\
					& & \textsc{APNEWS} & \textsc{IMDB} & \textsc{BNC} \\ \hline
					\multirow{6}{*}{50} & $\text{LDA}^{*}$  & 0.125 & 0.084 & 0.106 \\
					& $\text{NTM}^{*}$ & 0.075 & 0.064 & 0.081 \\
					& $\text{TDLM(s)}^{*}$& 0.149 & 0.104 & 0.102 \\
					& $\text{TDLM(l)}^{*}$& 0.130 & 0.088 & 0.095 \\
					& Topic-RNN(s) & 0.134 & 0.103 & 0.102 \\
					& Topic-RNN(l) & 0.127 & 0.096 & 0.100\\
					& TCNLM(s) & $\textbf{0.159}$ & $\textbf{0.106}$ &  \textbf{0.114}\\
					& TCNLM(l) & $0.152$& $0.100$& $0.101$\\ \hline
					\multirow{6}{*}{100} & $\text{LDA}^{*}$  & 0.136 & 0.092 & \textbf{0.119} \\
					& $\text{NTM}^{*}$ & 0.085 & 0.071 & 0.070 \\
					& $\text{TDLM(s)}^{*}$& 0.152 & 0.087 & 0.106 \\
					& $\text{TDLM(l)}^{*}$& 0.142 & 0.097 & 0.101 \\
					& Topic-RNN(s) &0.158 &0.096 & 0.108 \\
					& Topic-RNN(l) &0.143 &0.093 & 0.105\\
					& TCNLM(s) & $\textbf{0.160}$  & $\textbf{0.101}$ & $0.111$ \\
					& TCNLM(l) & $0.152$  & $0.098$& $0.104$\\ \hline 
					\multirow{6}{*}{150} & $\text{LDA}^{*}$  & 0.134 & 0.094 & \textbf{0.119} \\
					& $\text{NTM}^{*}$ & 0.078 & 0.075 & 0.072 \\
					& $\text{TDLM(s)}^{*}$& 0.147 & 0.085 & 0.100 \\
					& $\text{TDLM(l)}^{*}$& 0.145 & 0.091 & 0.104 \\
					& Topic-RNN(s) &0.146 & 0.089 & 0.102 \\
					& Topic-RNN(l) &0.137 & 0.092 & 0.097\\
					& TCNLM(s) & $0.153$ & \textbf{0.096} & $0.107$ \\
					& TCNLM(l) & $\textbf{0.155}$ & $0.093$ & $0.102$ \\
					\specialrule{.1em}{.05em}{.05em} 
			\end{tabular}}
			\label{Table:topic_coherence}
			\caption{Topic coherence scores of different models on \textsc{APNEWS}, \textsc{IMDB} and \textsc{BNC}. (s) and (l) indicate small and large model, respectively.($*$) taken from~\citet{lau2017topically}.}
		\end{center}
		\vspace{-2em}
	\end{table} 
	
	\begin{table*}[!htbp]
		\begin{center}
			\scalebox{0.6}{
				\begin{tabular}{l | c  l }
					\specialrule{.1em}{.05em}{.05em}
					\textbf{Data} &  \textbf{Topic} & \textbf{Generated Sentences} \\ \hline
					\multirow{8}{*}{\textsc{APNEWS}} & army & $\bullet$ a female sergeant, serving in the fort worth, has served as she served in the military in iraq .\\
					& animal & $\bullet$ most of the bear will have stumbled to the lake .\\
					& medical& $\bullet$ physicians seeking help in utah and the nih has had any solutions to using the policy and uses offline to be fitted with a testing or body .\\
					& market & $\bullet$ the company said it expects revenue of \$ $<$unk$>$ million to \$ $<$unk$>$ million in the third quarter .\\
					&lottory & $\bullet$ where the winning numbers drawn up for a mega ball was sold .\\ \cdashline{2-3}
					& \multirow{2}{*}{army+terrorism} & $\bullet$ the taliban 's presence has earned a degree from the 1950-53 korean war in pakistan 's historic life since 1964 , with two example of $<$unk$>$\\
					& &\quad soldiers from wounded iraqi army shootings and bahrain in the eastern army . \\ 
					& animal+lottory & $\bullet$ she told the newspaper that she was concerned that the buyer was in a neighborhood last year and had a gray wolf . \\ \hline
					\multirow{9}{*}{\textsc{IMDB}} & horror & $\bullet$ the killer is a guy who is n't even a zombie .\\
					& action & $\bullet$ the action is a bit too much , but the action is n't very good .\\
					& \multirow{2}{*}{family} & $\bullet$ the film is also the story of a young woman whose $<$unk$>$ and $<$unk$>$ and very yet ultimately sympathetic , $<$unk$>$ relationship , $<$unk$>$ ,\\
					& &  \quad and palestine being equal , and the old man , a $<$unk$>$ .\\
					& children & $\bullet$ i consider this movie to be a children 's film for kids .\\
					& war & $\bullet$ the documentary is a documentary about the war and the $<$unk$>$ of the war .\\ \cdashline{2-3}
					&horror+negative & $\bullet$ if this movie was indeed a horrible movie i think i will be better off the film . \\
					& \multirow{2}{*}{sci-fi+children} & $\bullet$ paul thinks him has to make up when the $<$unk$>$ eugene discovers defeat in order to take too much time without resorting to mortal bugs ,\\
					& & \quad and then finds his wife and boys . \\ \hline
					\multirow{7}{*}{\textsc{BNC}} & environment & $\bullet$ environmentalists immediate base calls to defend the world .\\
					& education & $\bullet$ the school has recently been founded by a $<$unk$>$ of the next generation for two years .\\
					& politics & $\bullet$ a new economy in which privatization was announced on july 4 .\\
					& business & $\bullet$ net earnings per share rose $<$unk$>$ \% to \$ $<$unk$>$ in the quarter , and \$ $<$unk$>$ m , on turnover that rose $<$unk$>$ \% to \$ $<$unk$>$ m. \\
					& facilities & $\bullet$ all rooms have excellent amenities .\\ \cdashline{2-3}
					& environment+politics & $\bullet$ the commission 's report on oct. 2 , 1990 , on jan. 7 denied the government 's grant to " the national level of water " . \\
					& art+crime & $\bullet$ as well as 36, he is returning freelance into the red army of drama where he has finally been struck for their premiere . \\
					\specialrule{.1em}{.05em}{.05em}
			\end{tabular}}
			\label{Table:generateSentences}
			\caption{Generated sentences from given topics. More examples are provided in the Supplementary Material. }
			\vspace{-1em}
		\end{center}
	\end{table*}

	\subsection{Topic Model Evaluation}
	
	We evaluate the topic model by inspecting the coherence of inferred topics~\citep{chang2009reading,newman2010automatic,mimno2011optimizing}.
	Following~\citet{lau2014machine}, we compute topic coherence using normalized PMI (NPMI). Given the top $n$ words of a topic, the coherence is calculated based on the sum of pairwise NPMI scores between topic words, where the word probabilities used in the NPMI calculation are based on co-occurrence statistics mined from English Wikipedia with a sliding window. In practice, we average topic coherence over the top $5/10/15/20$ topic words. To aggregate topic coherence score for a trained model, we then further average the coherence scores over topics. 
	For comparison, we use the following baseline topic models:
	\begin{itemize}
		\item \textbf{LDA}: LDA~\citep{blei2003latent} is used as a baseline topic model. We use LDA to learn the topic distributions for LDA+LSTM. 
		\item \textbf{NTM}: We evaluate the neural topic model proposed in~\citet{cao2015novel}. The document-topic and topic-words multinomials are expressed using neural networks. N-grams embeddings are incorporated as inputs of the model. 
		\item \textbf{TDLM}~\citep{lau2017topically}: The same model as used in the language model evaluation.
		\item \textbf{Topic-RNN}~\citep{dieng2016topicrnn}: The same model as used in the language model evaluation.
	\end{itemize}
	Results are summarized in Table 4.
	Our TCNLM achieves promising results. Specifically, (\textit{i}) we achieve the best coherence performance over \textsc{APNEWS} and \textsc{IMDB}, and are relatively competitive with LDA on \textsc{BNC}. (\textit{ii}) We also observe that a larger model may result in a slightly worse coherence performance. One possible explanation is that a larger language model may have more impact on the topic model, and the inherited stronger sequential information may be harmful to the coherence measurement. (\emph{iii}) Additionally, the advantage of our TCNLM over Topic-RNN indicates that our TCNLM supplies a more powerful topic guidance. 
	
	\begin{figure*}
		\centering
		\includegraphics[scale=0.4]{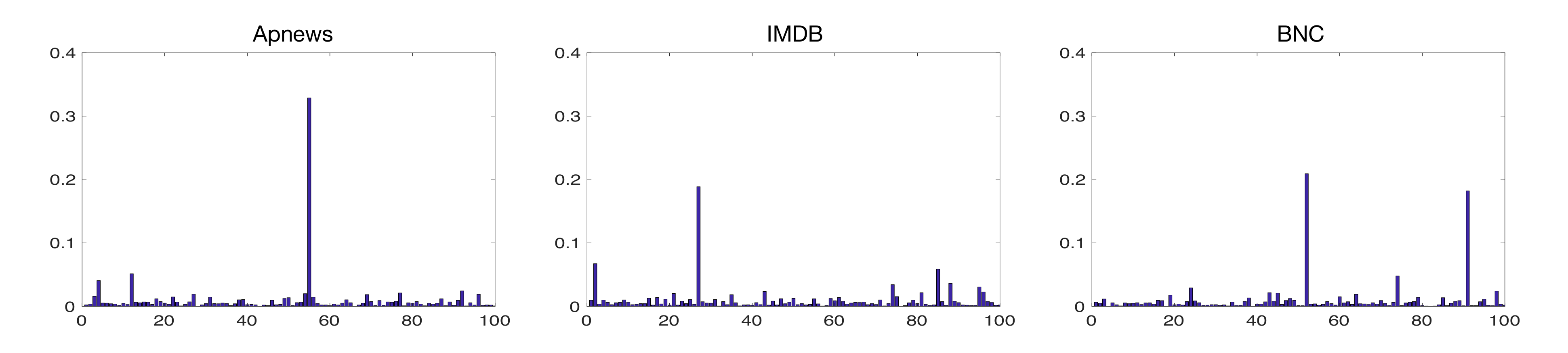}
		\label{figure: topicdist}
		\caption{Inferred topic distributions on one sample document in each dataset. Content of the three documents is provided in the Supplementary Mateiral.}
		\vspace{-1em}
	\end{figure*} 
	
	In order to better understand the topic model, we provide the top 5 words for 10 randomly chosen topics on each dataset (the boldface word is the topic name summarized by us), as shown in Table 3. These results correspond to the small network with 100 neurons. We also present some inferred topic distributions for several documents from our TCNLM in Figure 2. The topic usage for a specific document is sparse, demonstrating the effectiveness of our NTM. More inferred topic distribution examples are provided in the Supplementary Material. 
	
	\subsection{Sentence Generation}
	Another advantage of our TCNLM is its capacity to generate meaningful sentences conditioned on given topics. Given topic $\textit{i}$,  we construct an LSTM generator by using only the $i$-th factor of $\Wv_b$ and $\Uv_b$. Then we start from a zero hidden state, and greedily sample words until an end token occurs. Table 5 shows the generated sentences from our TCNLM learned with 50 topics using the small network. Most of the sentences are strongly correlated with the given topics. More interestingly, we can also generate reasonable sentences conditioned on a mixed combination of topics, even if the topic pairs are divergent, \textit{e.g.}, ``animal'' and ``lottory'' for \textsc{APNEWS}. More examples are provided in the Supplementary Material. It shows that our TCNLM is able to generate topic-related sentences, providing an interpretable way to understand the topic model and the language model simulaneously. These qualitative analysis further demonstrate that our model effectively assembles the meaning of topics to generate sentences. 
	
	\subsection{Empirical Comparison with Naive MoE}
	We explore the usage of a naive MoE language model as in (\ref{eq:MoE}). In order to fit the model on a single GPU machine, we train a NTM with $30$ topics and each NLM of the MoE is a 1-layer LSTM with $100$ hidden units. Results are summarized in Table 6.
	Both the naive MoE and our TCNLM provide better performance than the basic LSTM. Interestingly, though requiring less computational cost and storage usage, our TCNLM outperforms the naive MoE by a non-trivial margin. We attribute this boosted performance to the ``structure'' design of our matrix factorization method. The inherent topic-guided factor control significantly prevents overfitting, and yields efficient training, demonstrating the advantage of our model for transferring semantic knowledge learned from the topic model to the language model. 
	
	\begin{table}[!htbp]
		\begin{center}
			\scalebox{0.93}{
				\begin{tabular}{c |c | c | c  }
					\specialrule{.1em}{.05em}{.05em} 
					\textbf{Dataset} & \textbf{basic-LSTM} & \textbf{naive MoE} &\textbf{TCNLM} \\ \hline
					\textsc{APNEWS} & 101.62 & 85.87& \textbf{82.67} \\ \hline
					\textsc{IMDB} & 105.29 & 96.16 & \textbf{94.64} \\ \hline 
					\textsc{BNC} & 146.50 & 130.01 & \textbf{125.09}\\
					\specialrule{.1em}{.05em}{.05em} 
			\end{tabular}}
			\caption{Test perplexity comparison between the naive MoE implementation and our TCNLM on \textsc{APNEWS}, \textsc{IMDB} and \textsc{BNC}.}
			\label{Table:MoE Compare}
		\end{center}
		\vspace{-10pt}
	\end{table} 

	\section{Conclusion}
	\vspace{-10pt}
	
	We have presented Topic Compositional Neural Language Model (TCNLM), a new method to learn a topic model and a language model simultaneously. The topic model part captures the global semantic meaning in a document, while the language model part learns the local semantic and syntactic relationships between words. The inferred topic information is incorporated into the language model through a Mixture-of-Experts model design. Experiments conducted on three corpora validate the superiority of the proposed approach. Further, our model infers sensible topics, and has the capacity to generate meaningful sentences conditioned on given topics.
	One possible future direction is to extend the TCNLM to a conditional model and apply it for the machine translation task. 
	
	\clearpage
	\bibliographystyle{abbrvnat}
	\bibliography{topics}
	
	\clearpage
	\twocolumn[{%
		\centering
		\Large \bf{Supplementary Material for: }
		\\\bf{Topic Compositional Neural Language Model}
		\\[1.5em]
	}]
	
	\appendix
	
	\section{Detailed model inference}
	
	We provide the detailed derivation for the model inference. Start from~(\ref{eq:joint_likelihood}), we have
	{\footnotesize
		\begin{align}
		& \log p(y_{1:M}, \dv | \mu_0, \sigma_0^2, \betav) \nonumber\\
		&=\log  \int_\tv \frac{p(\tv)}{q(\tv | \dv)} q(\tv | \dv) p(\dv | \tv)   \prod_{m=1}^{M}   p(y_m | y_{1:m-1}, \tv)  d\tv ~\nonumber\\
		&= \log \mathbb{E}_{q(\tv | \dv)} \left( \frac{p(\tv)}{q(\tv | \dv)} p(\dv | \tv)   \prod_{m=1}^{M}   p(y_m | y_{1:m-1}, \tv) \right) ~\nonumber\\
		&\geq  \mathbb{E}_{q(\tv | \dv)}\left( \log p( \dv | \tv) - \log \frac{q(\tv | \dv )}{p(\tv)} + \sum_{m=1}^{M}\log p(y_m | y_{1:m-1}, \tv) \right) ~\nonumber\\
		&=  \underbrace{ \mathbb{E}_{q(\tv | \dv)} \left( \log p( \dv | \tv) \right)- \text{KL}\left( q(\tv | \dv) || p(\tv | \mu_0, \sigma_0^2) \right) }_{\text{neural topic model}} \nonumber\\
		&+  \underbrace{\mathbb{E}_{q(\tv | \dv)} \left( \sum_{n=1}^{M}\log p(y_m | y_{1:m-1}, \tv) \right)}_{\text{neural language model}} \nonumber \,.
		\end{align}
	}
	
	\section{Documents used to infer topic distributions}
	
	The documents used to infer the topic distributions ploted in Figure 2 are provided below. 
	
	\paragraph{Apnews}: \textit{colombia 's police director says six police officers have been killed and a seventh wounded in ambush in a rural southwestern area where leftist rebels operate . gen.       jose roberto leon tells the associated press that the officers were riding on four motorcycles when they were attacked with gunfire monday afternoon on a rural stret      ch of highway in the cauca state town of padilla .  he said a front of the revolutionary armed forces of colombia , or farc , operates in the area .  if the farc is r      esponsible , the deaths would bring to 15 the number of security force members killed since the government and rebels formally opened peace talks in norway on oct. 18       .  the talks to end a nearly five-decade-old conflict are set to begin in earnest in cuba on nov. 15 .}
	
	\paragraph{IMDB}: \textit{having just watched this movie for a second time , some years after my initial viewing , my feelings remain unchanged . this is a solid sci-fi drama that i enjoy very       much . what sci-fi elements there are , are primarily of added interest rather than the main substance of the film . what this movie is really about is wartime confl      ict , but in a sci-fi setting . it has a solid cast , from the ever reliable david warner to the up and coming freddie prinze jr , also including many british tv regu      lars ( that obviously add a touch of class :) , not forgetting the superb tcheky karyo .  i feel this is more of an ensemble piece than a starring vehicle .  reminisc      ent of wwii films based around submarine combat and air-combat ( the fighters seem like adaptations of wwii corsairs in their design , evoking a retro feel ) this is       one of few american films that i felt was not overwhelmed by sentiment or saccharine .  the sets and special effects are all well done , never detracting form the bel      ievability of the story , although the kilrathi themselves are rather under developed and one dimensional . this is a film more about humanity in conflict rather than       a film about exploring a new and original alien race or high-brow sci-fi concepts .  forget that it 's sci-fi , just watch and enjoy .}
	
	\textbf{BNC}: \textit{an army and civilian exercise went ahead in secret yesterday a casualty of the general election . the simulated disaster in exercise gryphon 's lift was a midair coll      ision between military and civilian planes over catterick garrison .  hamish lumsden , the ministry of defence 's chief press officer who arrived from london , said :       ' there 's an absolute ban on proactive pr during an election . '  journalists who reported to gaza barracks at 7.15 am as instructed were told they would not be all      owed to witness the exercise , which involved 24 airmobile brigade , north yorkshire police , fire and ambulance services , the county emergency planning department a      nd ' casualties ' from longlands college , middlesbrough .  the aim of the gryphon lift was to test army support for civil emergencies .  brief details supplied to th      e press outlined the disaster . a fully loaded civilian plane crashes in mid-air with an armed military plane over catterick garrison . the 1st battalion the green ho      wards and a bomb disposal squad cordon and clear an area scattered with armaments . 24 airmobile field ambulance , which served in the gulf war , tends a burning , pa      cked youth hostel hit by pieces of aircraft . 38 engineer regiment from claro barracks , ripon , search a lake where a light aircraft crashed when hit by flying wreck      age . civilian emergency services , including the police underwater team , were due to work alongside military personnel under the overall co-ordination of the police       .  mr lumsden said : ' there is a very very strict rule that during a general election nothing that the government does can intrude on the election process . '}
	
	\begin{figure*}
		\label{figure: topicdist_valid}
		\centering
		\includegraphics[scale=0.35]{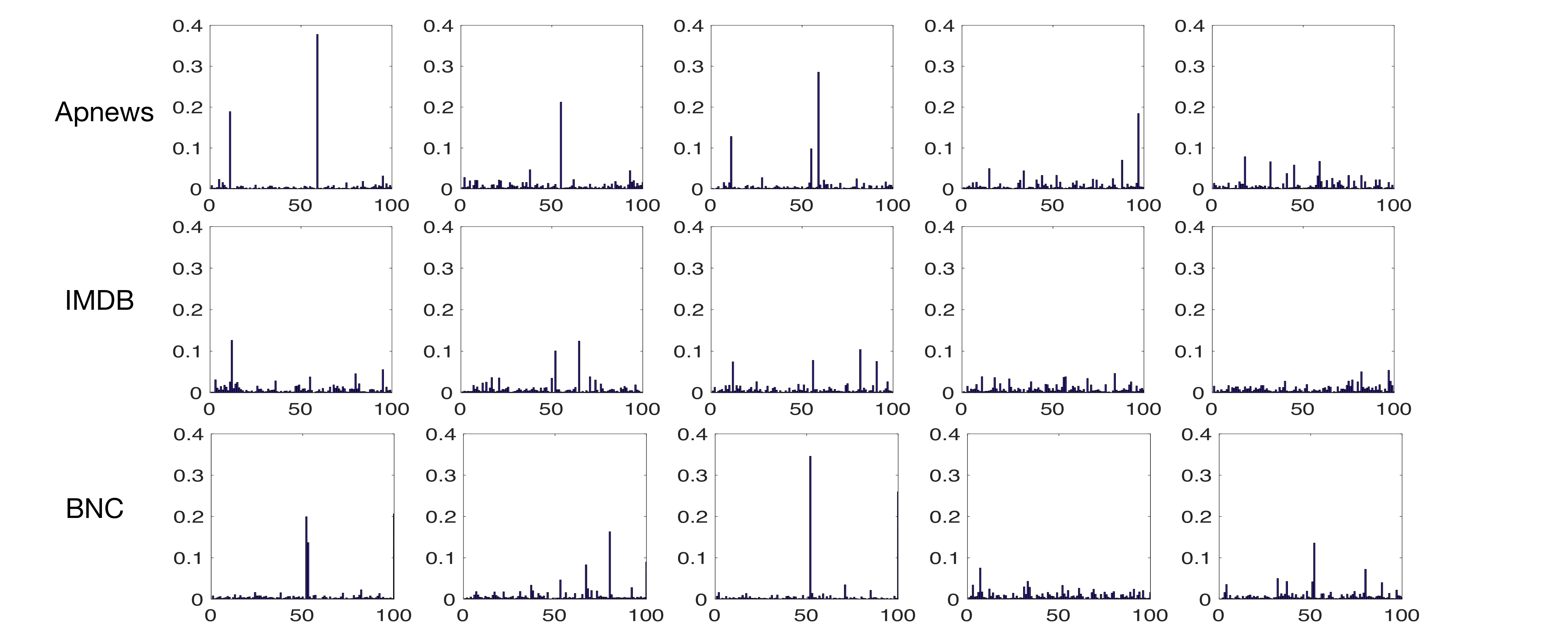}
		\caption{Inferred topic distributions for the first 5 documents in the development set over each dataset.}
	\end{figure*} 
	
	\section{More inferred topic distribution examples}
	We present the inferred topic distributions for the first 5 documents in the development set over each dataset in Figure 3.

	\section{More generated sentences}
	
	We present generated sentences using the topics listed in Table 3 for each dataset. The generated sentences for a single topic are provided in Table 7, 8, 9; the generated sentences for a mixed combination of topics are provided in Table 10.
	
	\begin{table*}
		\vspace{-1em}
		\begin{center}
			\scalebox{0.65}{
				\begin{tabular}{l |  l }
					\specialrule{.1em}{.05em}{.05em}
					\textbf{Topic} & \textbf{Generated Sentences} \\ \hline
					\multirow{5}{*}{army} & $\bullet$ a female sergeant, serving in the fort worth, has served as she served in the military in iraq .\\
					& $\bullet$ obama said the obama administration is seeking the state 's expected endorsement of a family by afghan soldiers  at the military\\ 
					& \quad  in world war ii, whose lives at the base of kandahar .\\
					& $\bullet$ the vfw announced final results on the u.s. and a total of \$ 5 million on the battlefield , but he 's still running for the democratic nomination \\
					&  \quad for the senate .\\ \hline
					\multirow{3}{*}{animal} & $\bullet$ most of the bear will have stumbled to the lake  .\\
					& $\bullet$ feral horses takes a unique mix to forage for their pets and is birth to humans .\\ 
					& $\bullet$ the zoo has estimated such a loss of half in captivity , which is soaked in a year .\\ \hline 
					\multirow{4}{*}{medical} & $\bullet$ physicians seeking help in utah and the nih has had any solutions to using the policy and uses offline to be fitted with a testing or body .\\
					& $\bullet$ that triggers monday for the study were found behind a list of breast cancer treatment until the study , as does in nationwide , has 60 days \\
					& \quad to sleep there .\\ 
					& $\bullet$ right stronger , including the virus is reflected in one type of drug now called in the clinical form of radiation .\\ \hline 
					\multirow{3}{*}{market} & $\bullet$ the company said it expects revenue of \$ $<$unk$>$ million to \$ $<$unk$>$ million in the third quarter  .\\
					& $\bullet$ biglari said outside parliament district of january, up \$ 4.30 to 100 cents per share , the last summer of its year to \$ 2 .\\ 
					& $\bullet$ four analysts surveyed by zacks expected \$ $<$unk$>$ billion .\\ \hline 
					\multirow{3}{*}{lottory} & $\bullet$ the numbers drawn friday night were $<$unk$>$  .\\
					& $\bullet$ where the winning numbers drawn up for a mega ball was sold .\\ 
					& $\bullet$ the jackpot is expected to be in july .\\ \hline 
					\multirow{4}{*}{terrorism} & $\bullet$ the russian officials have previously said the senior president made no threats .\\
					& $\bullet$ obama began halting control of the talks friday and last year in another round of the peace talks after the north 's artillery attack there \\
					& \quad wednesday have $<$unk$>$ their highest debate over his cultural weapons soon .\\
					& $\bullet$ the turkish regime is using militants into mercenaries abroad to take on gates was fired by the west and east jerusalem in recent years\\ \hline
					\multirow{3}{*}{law} & $\bullet$ the eeoc lawsuit says it 's entitled to work time for repeated reporting problems that would kick a nod on cheap steel from the owner \\
					& $\bullet$ the state allowed marathon to file employment , and the ncaa has a broken record of sale and fined for \$ $<$unk$>$ for a check\\
					& $\bullet$ the taxpayers in the lawsuit were legally alive and march $<$unk$>$ , or past at improper times of los alamos \\ \hline
					\multirow{4}{*}{art} & $\bullet$ quentin tarantino 's announcements that received the movie $<$unk$>$ spanish\\
					& $\bullet$ cathy johnson , jane 's short-lived singer steve dean and " the broadway music musical show , " the early show , " adds classics , $<$unk$>$ \\
					& \quad or 5,500 , while restaurants have picked other $<$unk$>$ next\\
					& $\bullet$ katie said , he 's never created the drama series : the movies could drops into his music lounge and knife away in a long forgotten gown \\ \hline
					\multirow{5}{*}{transportation} & $\bullet$ young and bernard madoff would pay more than \$ 11 million in banks , the airline said announced by the $<$unk$>$\\
					& $\bullet$ the fraud case included a delta 's former business travel business official whose fake cards " led to the scheme , " and to have been more \\
					& \quad than \$ 10,000 .\\
					& $\bullet$ a former u.s. attorney 's office cited in a fraud scheme involving two engines , including mining companies led to the government from \\
					& \quad the government .\\ \hline
					\multirow{5}{*}{education} & $\bullet$ the state 's $<$unk$>$ school board of education is not a $<$unk$>$ .\\
					& $\bullet$ assembly member $<$unk$>$, charter schools chairman who were born in new york who married districts making more than lifelong education \\
					& \quad play the issue , tells the same story that they 'll be able to support the legislation .\\
					& $\bullet$ the state 's leading school of grant staff has added the current schools to $<$unk$>$ students in a $<$unk$>$ class and ripley aims to serve \\
					& \quad child $<$unk$>$ and social sciences areas filled into in may and the latest resources\\ 
					\specialrule{.1em}{.05em}{.05em}
			\end{tabular}}
			\vspace{-0.5em}
			\label{Table:gen_apnews}
			\caption{More generated sentences using topics learned from \textsc{APNEWS}. }
		\end{center}
	\end{table*} 
	
	\begin{table*}[!htbp]
		\vspace{-1em}
		\begin{center}
			\scalebox{0.65}{
				\begin{tabular}{l |  l }
					\specialrule{.1em}{.05em}{.05em}
					\textbf{Topic} & \textbf{Generated Sentences} \\ \hline
					\multirow{4}{*}{horror} & $\bullet$ the killer is a guy who is n't even a zombie .\\
					& $\bullet$ the cannibals want to one scene , the girl has something out of the head and chopping a concert by a shark in the head , and he heads in the \\
					& \quad shower while somebody is sent to $<$unk$>$ .\\
					& $\bullet$ a bunch of teenage slasher types are treated to a girl who gets murdered by a group of friends .\\ \hline
					\multirow{4}{*}{action} & $\bullet$ the action is a bit too much , but the action is n't very good .\\
					& $\bullet$ some really may be that the war scene was a trademark $<$unk$>$ when fighting sequences were used by modern kung fu 's rubbish .\\
					& $\bullet$ action packed neatly , a fair amount of gunplay and science fiction acts as a legacy to a cross between 80s and , great gunplay and scenery .\\ \hline
					\multirow{5}{*}{family} & $\bullet$ the film is also the story of a young woman whose $<$unk$>$ and $<$unk$>$ and very yet ultimately sympathetic , $<$unk$>$ relationship , $<$unk$>$ .\\
					& \quad , and palestine being equal , and the old man , a $<$unk$>$ .\\
					& $\bullet$ catherine seeks work and her share of each other , a $<$unk$>$ desire , and submit to her , but he does not really want to rethink her issues , \\
					& \quad or where he aborted his mother 's quest to $<$unk$>$ .\\
					& $\bullet$ then i 'm about the love , but after a family meeting , her friend aditya ( tatum $<$unk$>$ ) marries a 16 year old girl , will be able to \\
					& \quad understand the amount of her boyfriend anytime\\ \hline
					\multirow{3}{*}{children} & $\bullet$ snoopy starts to learn what we do but i 'm still bringing up in there .\\
					& $\bullet$ i consider this movie to be a children 's film for kids .\\
					& $\bullet$ my favorite childhood is a touch that depicts her how the mother was what they apparently would 've brought it to the right place to fox .\\ \hline
					\multirow{3}{*}{war} & $\bullet$ the documentary is a documentary about the war and the $<$unk$>$ of the war. \\
					& $\bullet$ one of the major failings of the war is that the germans also struggle to overthrow the death of the muslims and the nazi regime , and $<$unk$>$ .\\
					& $\bullet$ the film goes , far as to the political , but the news that will be $<$unk$>$ at how these people can be reduced to a rescue .\\ \hline
					\multirow{5}{*}{detective} & $\bullet$ hopefully that 's starting $<$unk$>$ as half of rochester takes the character in jane 's way , though holmes managed to make tyrone power \\
					& \quad perfected a lot of magical stuff , playing the one with hamlet today .\\
					& $\bullet$ while the film was based on the stage adaptation , i know she looked up to suspect from the entire production .\\
					& $\bullet$ there was no previous version in my book i saw , only to those that read the novel , and i thought that no part that he was to why are far \\
					& \quad more professional .\\ \hline
					\multirow{3}{*}{sci-fi} & $\bullet$ the monster is much better than the alien in which the $<$unk$>$ was required for nearly every moment of the film .\\
					& $\bullet$ were the astronauts feel like enough to challenge the space godzilla , where it first prevails \\
					& $\bullet$ but the adventure that will arise from the earth is having a monster that can make it clear that the aliens are not wearing the $<$unk$>$ all that \\
					& \quad will $<$unk$>$ the laser .\\ \hline
					\multirow{3}{*}{negative} & $\bullet$ the movie reinforces my token bad ratings - it 's the worst movie i have ever seen .\\
					& $\bullet$ it was pretty bad , but aside from a show to the 2 idiots in their cast members , i 'm psychotic. \\
					& $\bullet$ we had the garbage using peckinpah 's movies with so many $<$unk$>$ , i can not recommend this film to anyone else .\\ \hline
					\multirow{3}{*}{ethic} & $\bullet$ englund earlier in a supporting role , a closeted gay gal reporter who apparently hopes to disgrace the girls in being sexual .\\
					& $\bullet$ this film is just plain stupid and insane , and a little bit of cheesy .\\
					& $\bullet$ the film is well made during its turbulent , exquisite , warm and sinister joys , while a glimpse of teen relationships .\\ \hline
					\multirow{4}{*}{episode} & $\bullet$ 3 episodes as a $<$unk$>$ won 3 emmy series !\\
					& $\bullet$ i remember the sopranos on tv , early 80 's , ( and in my opinion it was an abc show made to a minimum ) . .\\
					& $\bullet$ the show is notable ( more of course , not with the audience ) , and most of the actors are excellent and the overall dialogue is nice to \\
					& \quad watch ( the show may have been a great episode )\\
					\specialrule{.1em}{.05em}{.05em}
			\end{tabular}}
			\label{Table:gen_imdb}
			\vspace{-0.5em}
			\caption{More generated sentences using topics learned from \textsc{IMDB}. }
		\end{center}
	\end{table*}

	\begin{table*}[!htbp]
		\begin{center}
			\scalebox{0.7}{
				\begin{tabular}{l |  l }
					\specialrule{.1em}{.05em}{.05em}
					\textbf{Topic} & \textbf{Generated Sentences} \\ \hline
					\multirow{4}{*}{environment} & $\bullet$ environmentalists immediate base calls to defend the world .\\
					& $\bullet$ on the formation of the political and the federal space agency , the ec 's long-term interests were scheduled to assess global warming \\
					&\quad and aimed at developing programmes based on the american industrial plants .\\
					& $\bullet$ companies can reconstitute or use large synthetic oil or gas .\\ \hline
					\multirow{4}{*}{education} & $\bullet$ the school has recently been founded by a $<$unk$>$ of the next generation for two years .\\
					& $\bullet$ the institute for student education committees depends on the attention of the first received from the top of lasmo 's first round of \\
					& \quad the year .\\
					& $\bullet$ 66 years later the team joined blackpool , with $<$unk$>$ lives from the very good and beaten for all support crew -- and a new $<$unk$>$ \\
					&\quad student calling under ulster news may provide an $<$unk$>$ modern enterprise .\\ \hline
					\multirow{3}{*}{politics} & $\bullet$ the restoration of the government announced on nov. 4 that the republic's independence to direct cuba ( three years of electoral \\
					& \quad  $<$unk$>$ would be also held in the united kingdom ( $<$unk$>$ ) .\\
					& $\bullet$ a new economy in which privatization was announced on july 4 .\\
					& $\bullet$ agreements were hitherto accepted from simplified terms the following august 1969 in the april elections [ see pp. $<$unk$>$ . ] \\ \hline
					\multirow{3}{*}{business} & $\bullet$ net earnings per share rose $<$unk$>$ \% to \$ $<$unk$>$ in the quarter , and \$ $<$unk$>$ m , on turnover that rose $<$unk$>$ \% to \$ $<$unk$>$ m. \\
					& $\bullet$ the insurance management has issued the first quarter net profit up $<$unk$>$ - \$ $<$unk$>$ m , during turnover up $<$unk$>$ \% at \$ $<$unk$>$ \\
					&\quad m ; net profit for the six months has reported the cumulative losses of \$\\
					& $\bullet$ the first quarter would also have a small loss of five figures in effect , due following efforts of \$ 7.3 . \\ \hline
					\multirow{3}{*}{facilities} & $\bullet$ the hotel is situated in the $<$unk$>$ and the $<$unk$>$ .\\
					& $\bullet$ all rooms have excellent amenities .\\
					& $\bullet$ the restaurant is in a small garden , with its own views and $<$unk$>$ .\\ \hline
					\multirow{3}{*}{sports} & $\bullet$ the $<$unk$>$ , who had a $<$unk$>$ win over the $<$unk$>$ , was a $<$unk$>$ goal .\\
					& $\bullet$ harvey has been an thrashing goal for both $<$unk$>$ and the institutional striker fell gently on the play-off tip and , through in regular \\
					& \quad $<$unk$>$ to in the season .\\
					& $\bullet$ botham 's team took the fourth chance over a season without playing .\\ \hline
					\multirow{3}{*}{art} & $\bullet$ radio code said the band 's first album ' newest ' army club and $<$unk$>$ album .\\
					& $\bullet$ they have a $<$unk$>$ of the first album , ' run ' for a ' $<$unk$>$ ' , the orchestra , which includes the $<$unk$>$ , which is and the band 's life .\\
					& $\bullet$ nearly all this year 's album 's a sell-out tour ! \\ \hline
					\multirow{4}{*}{award} & $\bullet$ super french for them to meet winners $:$ the label $<$unk$>$ with just \# 50,000 should be paid for a very large size .\\
					& $\bullet$ a spokeswoman for $<$unk$>$ said : this may have been a matter of practice . \\
					& $\bullet$ female speaker they 'll start discussing music at their home and why decisions are celebrating children again , but their popularity has \\
					& \quad arisen for environmental research . \\ \hline
					\multirow{3}{*}{expression} & $\bullet$ roirbak stared at him , and the smile hovering .\\
					& $\bullet$ but they must have seen it in the great night , aunt , so you made the blush poor that fool .\\
					& $\bullet$ making her cry , it was fair .\\ \hline
					\multirow{4}{*}{crime} & $\bullet$ the prosecution say the case is not the same .\\
					& $\bullet$ the chief inspector supt michael $<$unk$>$ , across bristol road , and delivered on the site that the police had been accused to take him \\
					& \quad job because it is not above the legal services .\\
					& $\bullet$ she was near the same time as one of the eight men who died taken prisoner and subsequently stabbed , where she was hit away .\\
					\specialrule{.1em}{.05em}{.05em}
			\end{tabular}}
			\caption{More generated sentences using topics learned from \textsc{BNC}. }
			\label{Table:gen_bnc}
		\end{center}
	\end{table*} 
	
	\begin{table*}
		\vspace{-1em}
		\begin{center}
			\scalebox{0.65}{
				\begin{tabular}{l | l |  l }
					\specialrule{.1em}{.05em}{.05em}
					\textbf{Data} & \textbf{Topic} & \textbf{Generated Sentences} \\ \hline
					\multirow{8}{*}{\textsc{APNEWS}}&\multirow{4}{*}{army+terrorism} & $\bullet$ the taliban 's presence has earned a degree from the 1950-53 korean war in pakistan 's historic life since 1964 , with two \\ 
					&&\quad example of $<$unk$>$ soldiers from wounded iraqi army shootings and bahrain in the eastern army .\\
					&& $\bullet$ at the same level , it would eventually be given the government administration 's enhanced military force since the war .\\ 
					&& $\bullet$ the $<$unk$>$ previously blamed for the attacks in afghanistan , which now covers the afghan army , and the united nations will be \\
					&&\quad  a great opportunity to practice it .\\ \cline{2-3}
					&\multirow{4}{*}{animal+lottory} & $\bullet$when the numbers began , the u.s. fish and wildlife service unveiled a gambling measure by agreeing to acquire a permit by animal \\
					&&\quad protection staff after the previous permits became selected from the governor 's office .\\
					&& $\bullet$ she told the newspaper that she was concerned that the buyer was in a neighborhood last year and had a gray wolf .\\ 
					&& $\bullet$ the tippecanoe county historical society says it was n't selling any wolf hunts .\\ \hline
					\multirow{9}{*}{\textsc{IMDB}} & \multirow{4}{*}{horror+negative} & $\bullet$ if this movie was indeed a horrible movie i think i will be better off the film .\\
					&& $\bullet$ he starts talking to the woman when the person gets the town, she suddenly loses children for blood and it 's annoying to death \\
					&&\quad even though it is up to her fans and baby. \\
					&&$\bullet$  what 's really scary about this movie is it 's not that bad . \\ \cline{2-3}
					& \multirow{5}{*}{sci-fi+children} & $\bullet$ mystery inc is a lot of smoke , when a trivial , whiny girl $<$unk$>$ troy $<$unk$>$ and a woman gets attacked by the $<$unk$>$ captain ( \\
					&&\quad played by hurley ) .\\ 
					&& $\bullet$ paul thinks him has to make up when the $<$unk$>$ eugene discovers defeat in order to take too much time without resorting to \\
					&&\quad mortal bugs , and then finds his wife and boys .\\
					&&$\bullet$  the turtles are grown up to billy ( as he takes the rest of the fire ) and the scepter is a family and is dying . \\ \hline
					\multirow{8}{*}{\textsc{BNC}} & \multirow{5}{*}{environment+politics} & $\bullet$ $<$unk$>$ shallow water area complex in addition an international activity had been purchased to hit $<$unk$>$ tonnes of nuclear power \\
					&&\quad at the un plant in $<$unk$>$ , which had begun strike action to the people of southern countries .\\
					&& $\bullet$ the national energy minister , michael $<$unk$>$ of $<$unk$>$ , has given a " right " route to the united kingdom 's european parliament \\
					&&\quad , but to be passed by $<$unk$>$ , the first and fourth states .\\
					&&$\bullet$  the commission 's report on oct. 2 , 1990 , on jan. 7 denied the government 's grant to " the national level of water " .\\ \cline{2-3}
					& \multirow{3}{*}{art+crime} & $\bullet$ as well as 36 , he is returning freelance into the red army of drama where he has finally been struck for their premiere .\\ 
					&& $\bullet$ by alan donovan , two arrested by one guest of a star is supported by some teenage women for the queen .\\
					&&$\bullet$  after the talks , the record is already featuring shaun $<$unk$>$ 's play ' $<$unk$>$ ' in the quartet of the ira bomb .\\
					\specialrule{.1em}{.05em}{.05em}
			\end{tabular}}
			\vspace{-1em}
			\label{Table:gen_apnews_mix}
			\caption{More generated sentences using a mixed combination of topics.}
		\end{center}
	\end{table*} 
	
\end{document}